\documentclass[conference]{IEEEtran}
\usepackage{times}

\usepackage[numbers]{natbib}
\usepackage{multicol}
\usepackage[bookmarks=true]{hyperref}

\usepackage{amsmath,amssymb}
\usepackage{float}

\usepackage{graphicx}
\usepackage{comment}
\usepackage{color}  
\usepackage{xcolor}
\usepackage{amsfonts} 
\definecolor{agreen}{rgb}{0.25, 0.51, 0.20}
\usepackage{subfig}
\usepackage{algorithm}
\usepackage{algorithmic}
\usepackage{amsmath}
\usepackage{url}

\begin{document}

%
\title{Learning Memory-Based Control for Human-Scale Bipedal Locomotion}


\author{\authorblockN{Jonah Siekmann, Srikar Valluri, Jeremy Dao, Lorenzo Bermillo, Helei Duan, Alan Fern, and Jonathan Hurst}
\authorblockA{Collaborative Robotics and Intelligent Systems Institute\\
    Oregon State University\\
    Corvallis, Oregon 97331\\
    \emph{\{siekmanj, valluris, jdao, bermillf, duanh, afern, jhurst\}@oregonstate.edu}}
}

\maketitle

\begin{abstract}
Controlling a non-statically stable biped is a difficult problem largely due to the complex hybrid dynamics involved.  Recent work has demonstrated the effectiveness of reinforcement learning (RL) for simulation-based training of neural network controllers that successfully transfer to real bipeds.  The existing work, however, has primarily used simple memoryless network architectures, even though more sophisticated architectures, such as those including memory, often yield superior performance in other RL domains. In this work, we consider recurrent neural networks (RNNs) for sim-to-real biped locomotion, allowing for policies that learn to use internal memory to model important physical properties. We show that while RNNs are able to significantly outperform memoryless policies in simulation, they do not exhibit superior behavior on the real biped due to overfitting to the simulation physics unless trained using dynamics randomization to prevent overfitting; this leads to consistently better sim-to-real transfer. We also show that RNNs could use their learned memory states to perform online system identification by encoding parameters of the dynamics into memory.

\end{abstract}

\IEEEpeerreviewmaketitle

\section{Introduction}

Reinforcement learning has shown significant promise as a tool for solving complex control problems such as manipulation and legged locomotion. Recent work in transferring these trained controllers from simulation onto real robots has also enjoyed encouraging results \cite{peng2017sim2real, openai2019rubikscube, openai2018manipulation, xie2019iterative, tan2018simtoreal, hwangbo2019learning}, but many of these approaches use simple memoryless policy architectures, which could limit the extraction of useful information. Use of memory-based architectures has the potential to yield better performance in partially-observed domains, which has been observed in a variety of applications \cite{koul2018fsrnn, arras2017RNNpredictions, cechin2003state}. In this work, we demonstrate, for the first time, the application of learned, memory-based control to dynamic locomotion on the bipedal robot Cassie produced by Agility Robotics.

Memory-based controllers, such as recurrent neural networks (RNN), are a potentially powerful choice for solving highly dynamic nonlinear control problems due to their ability to infer important information about systems that is not directly observable \cite{peng2017sim2real, wierstra2010recurrent}. Though classical control methods \cite{zmp2004, westervelt2003hybrid, atrias2015} have made exceptional progress in bipedal control, these approaches often require disturbance observers to account for model inaccuracies \cite{kim2018disturbance,paine2015actuator} and are generally memory-based, containing some sort of hidden state which is updated in real-time. These can be seen as either predictive or history-compressing mechanisms, usually requiring tedious hand-tuning of gains. Conversely, RNNs have been shown to be able to infer the dynamics of systems thanks to the memory contained in their hidden state, effectively performing implicit system identification \cite{peng2017sim2real, Heess2015memorybased}, suggesting that they could fill the role of a disturbance observer and make memory-based learned control policies more robust than other, non-memory-based methods.

\begin{figure}[t!]
\centering
\includegraphics[scale=0.4]{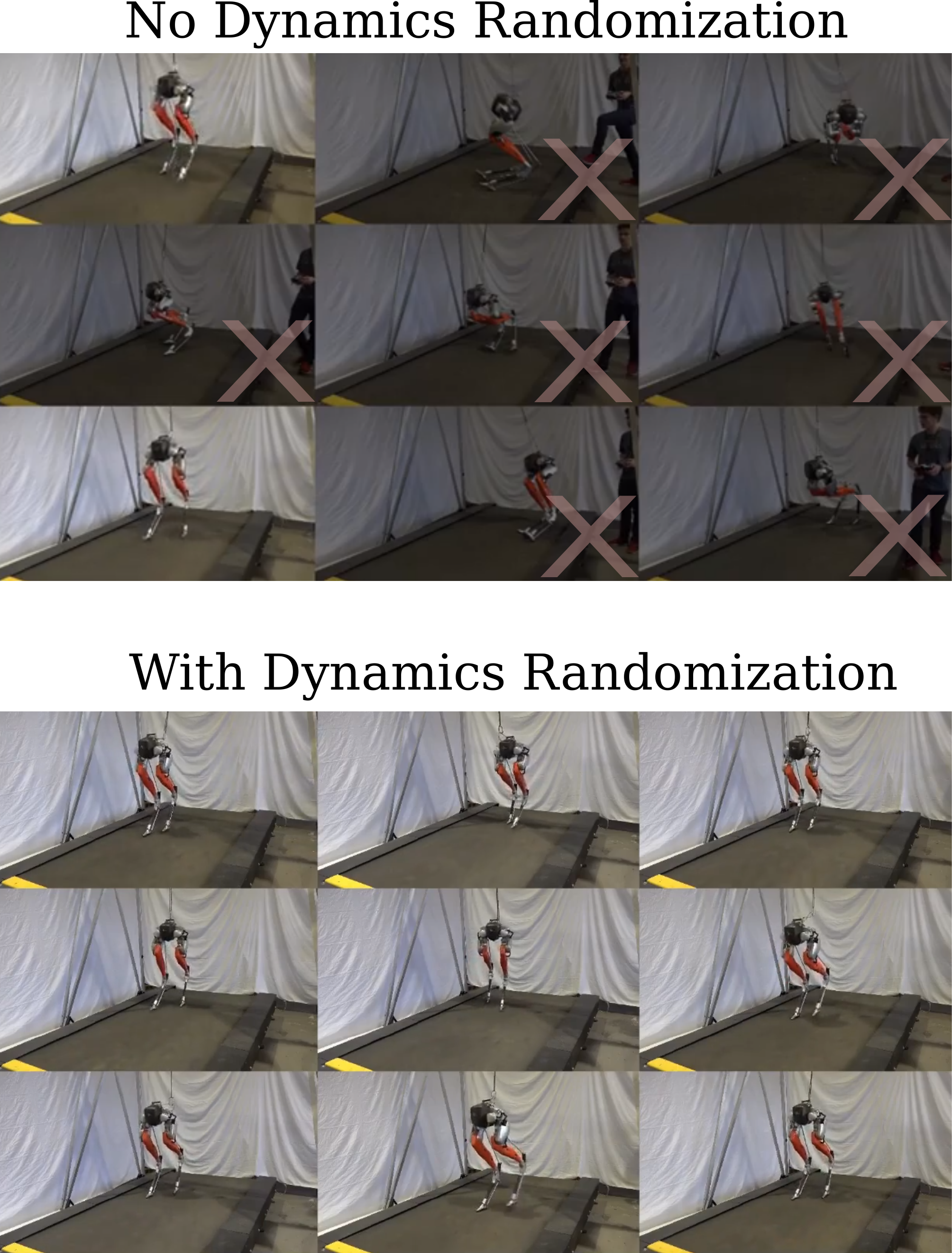}
\caption{We use recurrent neural networks and dynamics randomization to greatly improve the sim-to-real transfer rate and demonstrate learned, memory-based control on the bipedal robot, Cassie. RNNs trained without dynamics randomization are unable to consistently transfer to hardware (failures darkened and overlaid with X), while the same RNNs, when trained with dynamics randomization, are able to consistently transfer to the real world when trained with dynamics randomization.}
\label{fig:title_fig}
\end{figure}

While the representational power offered by RNNs has the potential to solve new classes of challenging control problems, it also increases the potential for overfitting. This can be problematic when training occurs in simulated environments, which do not necessarily correspond well to the real environment \cite{boeing2012leveraging}. In particular, an expressive RNN controller may learn to exploit details of the simulation dynamics that are maladaptive in the real world, leading to failure. 
As we demonstrate in our experiments, this is what we observe for our bipedal locomotion problem. That is, despite achieving significantly higher rewards than memoryless controllers in simulation, the same RNN controllers suffer from poor performance in the real world. 


A common way to help address this sim-to-real challenge is the use of dynamics randomization during simulation-based training. In particular, by training across different simulation environments with varying dynamics properties, the learned controller is less likely to exploit idiosyncrasies of any one dynamics setting. One of the main contributions of our work is to demonstrate that this approach is highly effective for training RNN controllers for the Cassie biped. We show that by randomizing a small number of dynamics parameters over reasonable ranges, the RNNs can be consistently trained in simulation and successfully transferred to the real world. We further give insight into the success of the RNN controllers, by demonstrating that the latent memory is able to effectively encode important dynamics parameters based on the observation history. 

\section{Background}

\subsection{Reinforcement Learning}

Reinforcement learning is a machine learning paradigm that seeks to train agents to maximize an expected reward through trial-and-error \cite{Sutton1998}. Reinforcement learning problems are often formalized as an agent interacting with a Markov decision process (MDP) in order to learn a behavior to maximize expected returns. This problem is typically presented in a manner in which an agent receives a state $s_t$ at timestep $t$ from the MDP, which the agent acts on based on its policy $\pi(a_t | s_t)$. The MDP's transition function receives the agent's action $a_t$ and returns the next state $s_{t+1}$ and reward $r_t$ based on the action taken. The policy is often a stochastic policy, in which case it is a function $\pi(a | s)$ which takes in a state $s$ and outputs the parameters of a distribution, usually the mean and standard deviation of a normal distribution. The reward $r = R(s, a)$ is a scalar signal that expresses how good a particular state-action pair is. The agent's goal is then to find an optimal policy that maximizes expected return $J(\pi)$, 
\begin{equation*}
    J(\pi) = \mathbb{E}_\pi \left[ \sum_{t=0}^{T}\gamma^{t} r_t \right]
\end{equation*}
where $T$ is the horizon of an episode, and $\gamma \in [0, 1]$ is the discount factor.

\subsection{Proximal Policy Optimization (PPO)}
An effective solution to many RL problems is the family of policy gradient algorithms, in which the gradient of the expected return with respect to the policy parameters is computed and used to update the parameters through gradient ascent.
PPO is a model-free policy gradient algorithm which samples data through interaction with the environment and optimizes a ``surrogate" objective function. PPO introduces a modified objective function that adopts clipped probability ratios which forms a pessimistic estimate of the policy's performance \cite{schulman2017proximal}. It also addresses the problem of excessive policy updates by restricting changes that move the probability ratio,
\begin{equation*}
r_t(\theta) = \frac{\pi_\theta(a_t | s_t)}{\pi_{\theta_{old}}(a_t | s_t)}
\end{equation*}
too far away from 1. The probability ratio is a measure of how different the current policy is from the previous policy (the policy before the last update). The smaller the ratio the greater the difference. The ``surrogate" objective function is then modified into the clipped objective:
\begin{equation*}
    L(\theta) = \mathbb{E}_t\left[min\left(r_t(\theta)\hat{A_t}, clip(r_t(\theta), 1-\epsilon, 1+\epsilon)\hat{A_t}\right)\right]
\end{equation*}
where $\epsilon$ is a tunable hyperparameter that increases or decreases the bounds which constrain the probability ratio $r_t(\theta)$. Clipping the probability ratio discourages the policy from changing too much and taking the minimum results in using the lower, pessimistic bound of the unclipped objective. Thus any change in the probability ratio $r_t(\theta)$ is included when it makes the objective worse, and otherwise is ignored \cite{schulman2017proximal}. This can prevent the policy from changing too quickly and leads to more stable learning. 

\subsection{RNNs for Control Problems}

RNNs have been successfully applied to many control domains using reinforcement learning, often resulting in performance superior to feedforward networks. \citet{UAVmemory} shows that using deep recurrent Q networks (DRQN) instead of conventional feedforward Q Networks in UAV navigation results in less collisions and more energy-efficient performance \cite{UAVmemory}. \citet{recurrentDQN} also use DRQNs for Atari games, finding that DRQNs with a single observation are a viable alternative to DQNs with a history of states, and that DRQNs are more robust to partial observability that non-memory based agents. It has been shown that RNNs can store and recollect information for arbitrarily long amounts of time \cite{schmidhuber1997lstm}, as well as perform system identification as noted in \citet{Heess2015memorybased} and further explored in \citet{peng2017sim2real}. Furthermore, RNNs can do so through gradient descent, without hand-tuning of hyperparameters, in contrast to feedforward networks which require access to hand-picked fixed window of state histories \cite{yu2017sysid}. 

Thus, while RNNs have been shown to be extremely successful in achieving superior performance to feedforward networks on a variety of robotic control tasks, some even on hardware \cite{peng2017sim2real}, they have not yet been demonstrated for the task of real-world robotic bipedal locomotion. This task differs significantly from other control tasks, such as robotic arm manipulation, due to the significant underactuation of the system and complicated contact dynamics.

\subsection{Dynamics Randomization}
Dynamics randomization \cite{peng2017sim2real} \cite{tan2018simtoreal} is the practice of randomizing physics parameters of the simulated environment in the hopes that training agents on a variety of possible dynamics will lead to better performance in the real world. \citet{tan2018simtoreal} leverage this technique to learn quadruped locomotion from scratch in a physics simulator and then deploy the learned controller into hardware, while \citet{peng2017sim2real} use a similar system to train a robotic arm to manipulate objects in the real world. They improve the robustness of control policies by simulating latency as well as physical properties such as mass, joint center of mass, joint damping, and other similar parameters of the environment.

More formally, using notation from \citet{peng2017sim2real}, the objective is to train a memory-based agent to perform manipulation tasks under the conditions set by the real world dynamics $p^*(s_{t+1}|s_t,a_t)$. However, sampling from these dynamics is not very time-efficient. Instead, the agent is trained across a wide range of possible dynamics by using a set of dynamics parameters $\mu$, sampled from a multivariate uniform distribution, to parameterize the simulation dynamics $\hat{p}_\mu(s_{t+1}|s_t,a_t)$, so the objective is reframed as attempting to maximize the expected return over the distribution of dynamics parameters $\rho_\mu$.

\section{Method}
\begin{figure}[t!]
\includegraphics[scale=1.45]{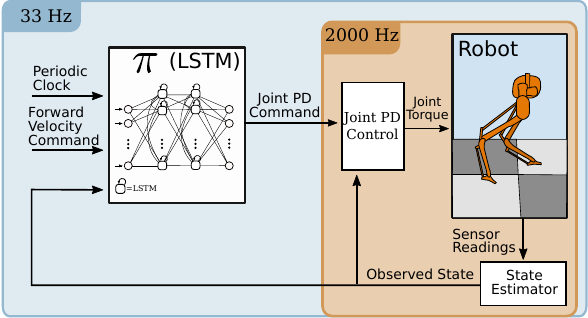}
\caption{We provide an RNN with a clock input, a velocity command, and information about the robot's state. The RNN produces joint position commands at 33Hz, which are translated into torque-level commands by a PD controller at 2kHz.}
\label{fig:control_diagram}
\end{figure}
\begin{figure*}[!htbp]
\centering
\includegraphics[scale=0.65]{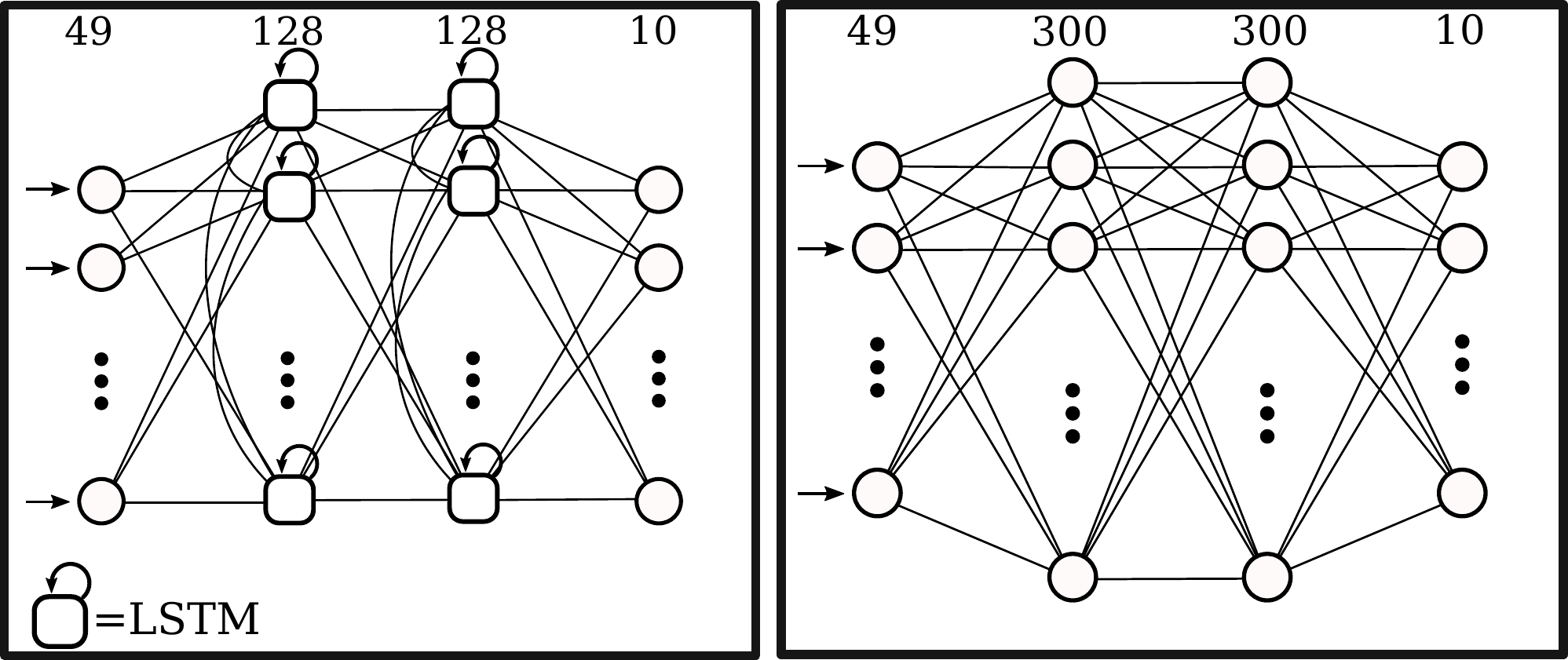}
\caption{A diagram of the RNN structure (left) and conventional NN structure (right) we use in our experiments. The recurrent policy has connections which loop back onto itself, so that information from previous timesteps is available. This produces a memory-like mechanism, allowing the RNN to encode things about the state history which may be useful for choosing actions. Both networks have approximately the same number of parameters.}
\end{figure*}

\subsection{State Space and Action Space}

The policy's input consists of:
\[ X_t =  \left\{
\begin{array}{ll}
      \smallskip
      f_{vel} & \text{desired forward speed} \\
      \smallskip
      \sin(\frac{2 \pi \omega}{L}) &  \text{clock input} \\
      \smallskip
      \cos(\frac{2 \pi \omega}{L}) & \text{clock input} \\
      \hat{q},\ \hat{\dot{q}} & \text{robot state} 
\end{array} 
\right. \]

Where $f_{vel}$ is the desired forward speed, $\sin(\frac{2 \pi \omega}{L})$ and $\cos(\frac{2 \pi \omega}{L})$ are clock inputs, $\omega$ is an integer that increments by one every policy evaluation step, $L$ is the phase length of the reference motion, and $\hat{q},\ \hat{\dot{q}}$ are a collection of sensor-based estimates of robot orientation, joint positions, and joint velocities. 
In total, the input dimension of the networks is 49. The outputs of the policy are simply raw motor PD targets, much like \citet{xie2019iterative}. 

The policy is evaluated every 30 milliseconds, or roughly 33Hz, while the PD controller operates at 2kHz, as can be seen in Figure \ref{fig:control_diagram}. The control rate was chosen because it corresponds to an even 60 cycles of the PD controller. The output of the policy is added to a constant ``neutral position" offset and then fed directly into the motor PD controllers. The ``neutral position" corresponds to the static motor positions of the robot standing at rest. For more details of the system setup we refer readers to \citet{xie2019iterative}, who use a similar system.

\subsection{Reward Design}
Our learning process makes use of a reference trajectory produced by an expert walking controller to help the policy learn in the initial stages of training. The reference trajectory contains a sequence of states of the robot walking forward at 1 m/s. Though the reference contains the full robot state at each timestep, we only use the center of mass position, orientation, and velocity, motor positions, and spring positions. All policies were trained to maximize the following reward function:

\begin{equation}
\begin{aligned} 
R=\displaystyle & 0.20  \cdot \text{exp}(-q_{\text{err}}) & + & 0.20  \cdot \text{exp}(-\dot{x}_{\text{err}}) \\
+ & 0.05 \cdot \text{exp}(-x_{\text{err}}) & + & 0.20  \cdot \text{exp}(-\dot{y}_{\text{err}}) \\ 
+ & 0.30 \cdot \text{exp}(-\text{orient}_{\text{err}})  & + & 0.05 \cdot \text{exp}(-\text{spring}_{\text{err}})
\end{aligned}
\end{equation}
$q_\text{err}$, $\dot x_\text{err}$, $x_\text{err}$, and $\text{spring}_\text{err}$ are squared error terms between the robot's joint positions, forward velocity, forward position, and passive spring position and that of the corresponding reference state. These terms encourage the policy to follow the reference motion. To prevent long-term sideways drift, we penalize the y velocity $\dot y_\text{err}$. To keep the robot facing straight, we use $\text{orientation}_\text{err}$ which is the quaternion difference between the robot's orientation and an orientation which faces straight ahead. 

It is important to note that while the reward function is partially based on a reference motion, very similarly to \citet{xie2019iterative}, the agent does not actually receive any information about the reference motion in the input space, aside from a clock input which corresponds roughly to the walking cycle of the reference motion. Though we believe that the recurrent policies do not have any theoretical reason for needing a clock input, we were not able to train any to walk on hardware without it.

\subsection{Dynamics Randomization}

At the start of each episode during training, we randomize 61 dynamics parameters. A description of these parameters can be found in Table \ref{table:dynamicsrand}.

\begin{table}[!h]
\centering
\begin{tabular}{|l|l|l|}
\hline
Parameter & Unit & Range \\ \hline
 Joint damping                      & Nms/rad   &   $[0.5, 1.5] \times \text{default values}$ \\ \hline
 Joint mass                         & kg        &   $[0.7, 1.3] \times \text{default values}$     \\ \hline
 Pelvis CoM (x)                     & cm        &   $[-25, 6]$ from origin \\ \hline
 Pelvis CoM (y)                     & cm        &   $[-7, 7]$ plus default  \\ \hline
 Pelvis CoM (z)                     & cm        &   $[-4, 4]$ plus default  \\ \hline

\end{tabular}
\caption{We chose ranges for each parameter based on estimates of uncertainty. In addition, when robot behaviors seemed sensitive to a particular parameter and this parameter seemed to be wrong, we increased the randomization range to reduce sensitivity.}
\label{table:dynamicsrand}
\end{table}

We aggressively randomize the pelvis center of mass because we believe the robot's center of mass in simulation differs significantly from the center of mass we see on hardware; the results we see on hardware appear to confirm this phenomenon. While this is not as principled as randomizing the entire robot's center of mass, we find that it introduces enough noise into the simulation dynamics to overcome the sim-to-real gap.

\subsection{Recurrent Proximal Policy Optimization}
We trained all policies with PPO, a model-free reinforcement learning algorithm. However, correctly calculating the gradient of an RNN requires the use of the backpropagation through time (BPTT) algorithm, which necessitates special measures when sampling from the replay buffer. Thus, we use PPO with the exception that for recurrent policies, instead of sampling individual timesteps from the replay buffer, we sample batches of entire trajectories. In addition, we collect statistics including the mean state $s_\mu$ and standard deviation $s_\sigma$ before training, and use these to normalize states during training and during evaluation. We found this prenormalization step to be very important for consistent performance. In addition, we used a threshold to stop the KL divergence between the old policy and new policy from growing too large by aborting the optimization step early if the threshold was surpassed. Pseudocode of the recurrent version of the algorithm that we used can be seen in Algorithm \ref{algo:rppo}.

\begin{algorithm}[]
\caption{Recurrent PPO}
\label{algo:rppo}
\begin{algorithmic}
\REQUIRE $kl_{thresh} \geq 0 \in \mathbb{R}$
\REQUIRE $N > 0 \in \mathbb{Z}, K > 0 \in \mathbb{Z}$
    \STATE Collect state mean $s_{\mu}$ and standard deviation $s_{\sigma}$
    \FOR{iteration=1,2,... }
        \STATE buff = []
        \STATE $\theta_{old} \leftarrow \theta$
        \item Collect $N$ rollouts from environment into buff. 
        \item Compute advantage estimates $\hat{A}_{\text{buff}}$.
        \FOR{epoch=1,2, ... $K$}
            \FOR{batch=1,2, ...}
                \STATE $\hat{s}, \hat{a}, \hat{A} \leftarrow $ sample trajectories from buff.
                \STATE Optimize surrogate $L$ wrt $\theta$, using $\hat{s}, \hat{a}, \hat{A}$
                \IF{$KL(\pi_\theta(\hat{s}, \hat{a}), \pi_{\theta_{old}}(\hat{s}, \hat{a})) > kl_{thresh}$}
                    \item break
                \ENDIF
            \ENDFOR
        \ENDFOR
    \ENDFOR
\end{algorithmic}
\end{algorithm}

\subsection{Network Architecture}

All policies had an input dimension of length 49, and an output dimension of size 10. The recurrent policy used was a Long Short-Term Memory \cite{schmidhuber1997lstm} network consisting of two LSTM layers of 128 units each, and a linear output layer. Our conventional neural network, or feedforward (FF), policies had two hidden layers of 300 units each, which was roughly equivalent in terms of parameter count to the recurrent policy. During training, we used a fixed action standard deviation with a value of $e^{-2}$.

The recurrent critic was also an LSTM network with two layers of 128 units each, with a single output representing the value function. The feedforward critic was similarly a network with two hidden layers of 300 units each, in order to maintain parameter-count equivalence with the recurrent critic. No information about the dynamics disturbances was provided to either the policy or the critic in the input space.

\section{Results}

\subsection{Simulation}
\begin{figure}[!h]
\centering
\includegraphics[scale=0.5]{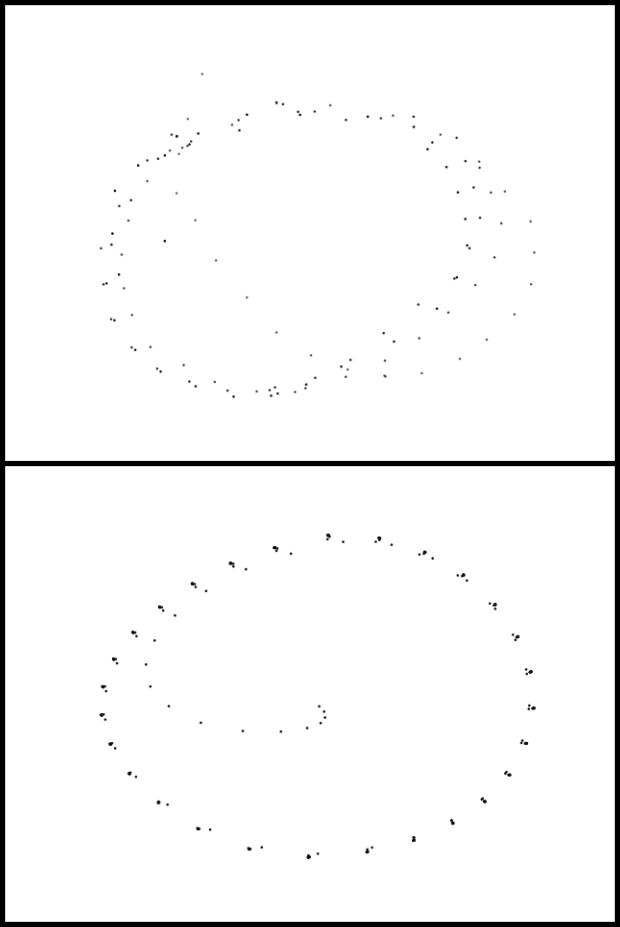}
\caption{A PCA projection of the hidden layer activations of a feedforward policy (top) and of an LSTM policy (bottom) could imply that using an LSTM policy better enables learning of cyclic behaviors. Each dot represents a two-dimensional projection of all latent states for a single timestep. Dots become lighter as a function of time.}
\label{fig:pca}
\end{figure}

\begin{figure}[!h]
\includegraphics[scale=0.57]{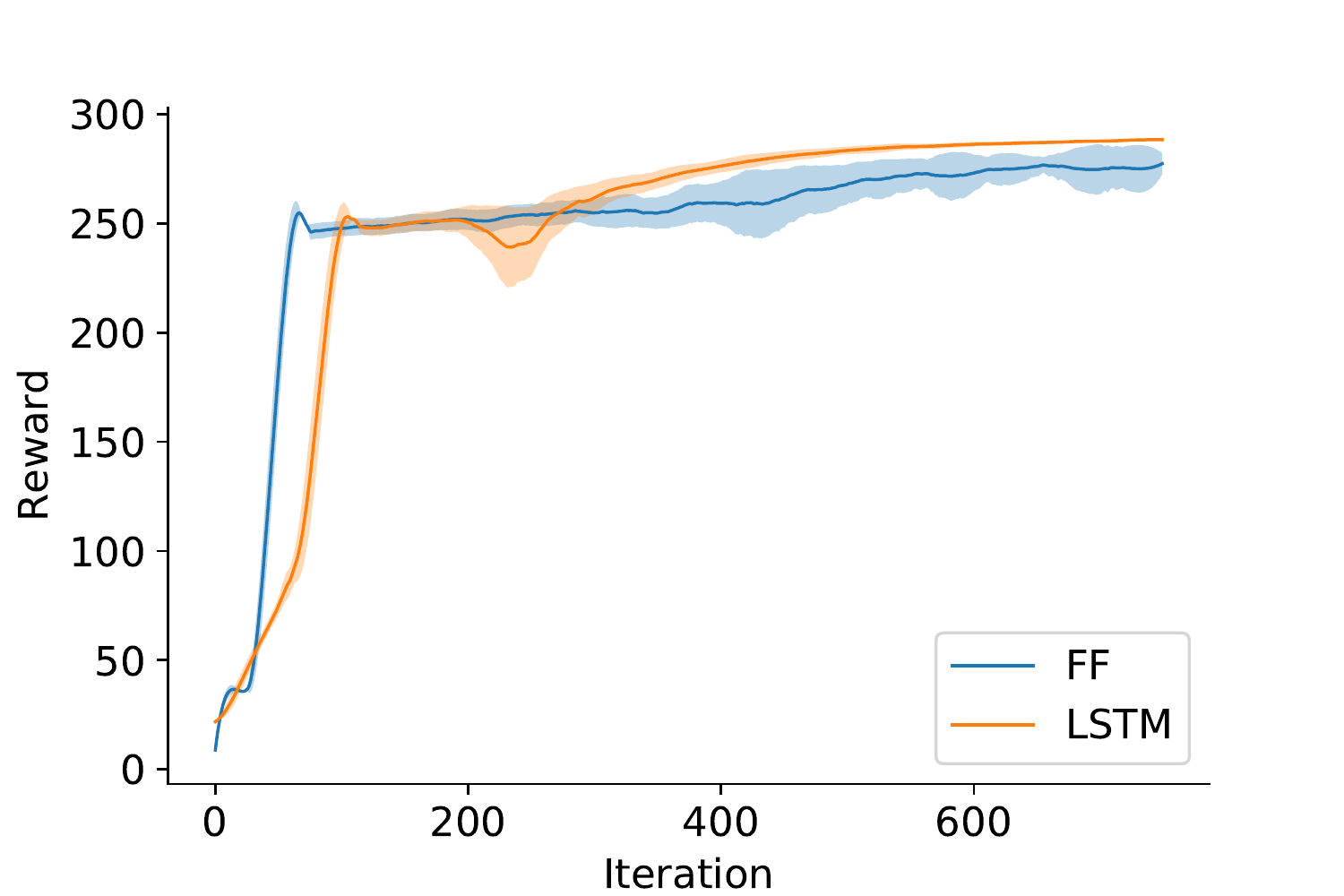}
\caption{Reward curve of LSTM and FF networks during training without dynamics randomization. The LSTM achieves a much higher reward with remarkably little variance, but both networks perform roughly the same on hardware.}
\label{fig:reward_curve)}
\end{figure}

We trained ten LSTM networks and ten FF networks with dynamics randomization, and ten LSTM networks and ten FF networks without dynamics randomization, each with separate random seeds. When training the recurrent policies, we used a batch size of 64 trajectories and a maximum trajectory length of 300 timesteps, equal to 9 seconds of simulation time. When training the feedforward policies, we used a batch size of 1024 timesteps. Each recurrent network took about sixteen hours to train on a machine with 56 CPU cores and 112 threads, while feedforward networks took about six hours on the same machine. All networks were trained for fifty million simulation timesteps, and each iteration we sampled about fifty thousand timesteps from the simulated environment. We used PyTorch \cite{pytorch} to train and execute the policy, and Adam \cite{kingma2014adam} as our optimizer. As can be seen in Figure \ref{fig:reward_curve)}, when trained without dynamics randomization, LSTM networks attain a significantly higher reward than feedforward networks, with surprisingly little variance. Feedforward networks obtain a notably lower reward, with high variance. We attribute this difference to the LSTM overfitting the dynamics of the simulation, as feedforward networks appear to perform roughly at the same level as LSTM networks in sim-to-real when both are trained without dynamics randomization.

We conducted a robustness test in simulation across ten chosen sets of dynamics, taken from the range in Table \ref{table:dynamicsrand}. The results can be seen in Table \ref{table:simulationdisturb}. We randomly sample ten sets of dynamics parameters and measure the average time each type of policy is able to remain upright and walking for each of the ten sets. As can be seen, dynamics randomization improves performance of both policy types and LSTM with dynamics randomization performs the best.

We also include a dimension-reduced visualization of the latent spaces of both a feedforward and LSTM policy using principal component analysis, seen in Figure \ref{fig:pca}. The projection of the latent states is in both cases highly cyclic, though the LSTM policy is much more neatly segmented into 28 distinct points, matching the clock cycle of the reference trajectory. The shape of these projections aligns with our intuition that a bipedal walking controller should learn behaviors which are cyclical, and could indicate that LSTM networks have an advantage in emulating this.

\begin{table}[!h]
\centering
\begin{tabular}{|c|r|r|r|r|}
\hline
\multicolumn{1}{|c} {Parameter Set} & \multicolumn{1}{|c} {LSTM} & \multicolumn{1}{|c} {LSTM DR} & \multicolumn{1}{|c} {FF} & \multicolumn{1}{|c|} {FF DR} \\ \hline
$\mu_1$     & 1.3s      & $>40.0$s  & 1.6s    & 17.1s     \\ \hline
$\mu_2$     & $>40.0$s    & $>40.0$s  & $>40$s  & $>40.0$s    \\ \hline
$\mu_3$     & 1.9s      & $>40.0$s  & 6.7s    & 14.3s      \\ \hline
$\mu_4$     & 1.5s      & $>40.0$s  & 1.6s    & 1.93s      \\ \hline
$\mu_5$     & $>40.0$s    & $>40.0$s  & 36.7s   & $>40.0$s    \\ \hline
$\mu_6$     & 1.6s      & $>40.0$s  & 1.7s    & 35.2s     \\ \hline
$\mu_7$     & $>40.0$s    & $>40.0$s  & 36.3s   & 13.7s     \\ \hline
$\mu_8$     & $>40.0$s    & $>40.0$s  & 4.8s    & 19.2s     \\ \hline
$\mu_9$     & 28.8s     & $>40.0$s  & 12.2s  & 23.6s      \\ \hline
$\mu_{10}$  & $>40.0$s    & $>40.0$s  & 6.7s  & 15.8s     \\ \hline
\textbf{Avg.}    & \textbf{23.5s}    & \textbf{40.0s}     & \textbf{14.8s}    & \textbf{22.1s}     \\ \hline
\end{tabular}
\caption{Average time (in seconds) that ten randomly seeded policies were able to walk in simulation subject to the conditions of the randomly chosen dynamics parameters $\mu_i$, sampled from the distributions specified in Table \ref{table:dynamicsrand}. The last row gives the averaged values over all 10 sets of dynamics parameters.}
\label{table:simulationdisturb}
\end{table}

\subsection{Hardware}

Results are best seen in our accompanying video \footnote{\url{https://youtu.be/V8_JVvdJt_I}}. All of the recurrent policies trained with dynamics randomization are able to walk forward for at least forty seconds on the real robot, while LSTMs and FF networks trained without randomization fail to consistently transfer to hardware.

We were unable to get FF policies trained with dynamics randomization to work on hardware, as they exhibited extremely shaky behavior and evaluating them would have likely resulted in damage to our robot. We believe that the reason for this shakiness is the aggressive amount of dynamics randomization we do in simulation, which interferes with the ability of memoryless agents to learn.

We note that all of the LSTM policies trained with dynamics randomization are able to successfully walk on Cassie, while only two LSTM policies trained without randomization are able to enter a stable walking gait. Feedforward networks perform slightly better than LSTM networks when both are trained without dynamics randomization, though much like \citet{xie2019iterative}, we observe significant shakiness and instability on FF networks, which is not present in the LSTM networks.

\begin{table}[!h]
\centering
\begin{tabular}{|c|r|r|r|}
\hline
\multicolumn{1}{|c}{Seed} & \multicolumn{1}{|c} {LSTM} & \multicolumn{1}{|c}{LSTM DR} & \multicolumn{1}{|c|}{FF} \\ \hline 
 1 & 7s     &$>40$s & 0s      \\ \hline
 2 & 7s     &$>40$s & 16s     \\ \hline
 3 & $>40$s &$>40$s & $>40s$  \\ \hline
 4 & 11s    &$>40$s & 9s  \\ \hline
 5 & 3s     &$>40$s & 33s  \\ \hline
 6 & 3s     &$>40$s & 0s  \\ \hline
 7 & 2s     &$>40$s & 5s \\ \hline
 8 & 0s     &$>40$s & 3s  \\ \hline
 9 & $>40$s &$>40$s & 7s  \\ \hline
 10 & 4s    &$>40$s & 10s  \\ \hline
\textbf{Avg.} & \textbf{11.7s} & \textbf{40s}   & \textbf{12.3s}  \\ \hline
\end{tabular}
\caption{Time (in seconds) that a policy resulting from a seed was able to walk in the real world. We cut off experiments that run for more than 40 seconds.}
\label{table:uptime}
\end{table}

Of particular difficulty to most policies was stepping in-place. We have reason to believe that the robot's true center of mass position is significantly different from the center of mass position we simulate. As a means of addressing this discrepancy, we randomize the pelvis center of mass x location in order to move the center of mass of the full robot forward and backward, exposing agents to potential differences between simulation and the real world. None of the policies trained without dynamics randomization were able to step in-place, though some were able to walk forward. This is likely because the robot center of mass discrepancy is made worse by the fact that low-speed or no-speed walking is much more dynamically unstable than walking at speed. All of the recurrent policies trained with dynamics randomization were able to transfer to hardware successfully and walk in a straight line for over forty seconds.

\subsection{Dynamics extraction}
We assume that in order for a memory-based agent to succeed in an environment with unknown dynamics, it is advantageous to maintain a compressed history of states that includes encoded estimates about those dynamics to determine which action to take. We hypothesize that it should be possible to extract these embedded dynamics from an RNN trained with dynamics randomization by imagining the hidden states of the RNN as the latent representation of an autoencoder, and use supervised learning to decode these latent representations back into the original dynamics.

Using these hidden state and ground truth dynamics pairs, we constructed datasets of 40,000 training data points, and 10,000 testing data points for each policy tested. We then trained a decoder network to predict the dynamics from the recurrent hidden states.

For most of the dynamics parameters, we found that dynamics randomization did not provide an advantage in allowing LSTMs to encode information about the dynamics into their hidden states. This is probably due to the state-history compression ability of RNNs, which contained enough information for the decoder network to make approximate guesses at the true value, regardless of whether the encoder had been trained with dynamics randomization. 

However, we were able to consistently predict the position of the pelvis center of mass in the sagittal axis with higher fidelity when it was encoded using LSTMs trained with dynamics randomization, as can be seen in Table \ref{table:com_error}. This is consistent with our expectations, as it was the dynamics parameter that we most aggressively randomized. This suggests that in simulation, joint damping and joint mass were simply not important enough to bother accounting for individually, and instead policies just learned to be as robust as possible to variations in these quantities instead of attempting to infer information about their actual values. Since we chose such a large range for pelvis center of mass, however, policies were unable to merely learn behavior that was robust across the entire range of possible values.

\begin{table}[!h]
\begin{tabular}{|l|r|r|}
\hline
Policy & Mean Absolute Error & Mean Percent Error\\ \hline
LSTM (randomization) & 1.77mm & 30.8\% \\ \hline 
LSTM (no randomization) & 2.99mm & 62.4\%  \\ \hline
\end{tabular}
\caption{Mean Absolute Error and Mean Percent Error for the decoder network's prediction of the pelvis center of mass in the sagittal axis. Here, the decoder networks are able to decode far more information about the pelvis center of mass from the memory of the policies trained with dynamics randomization than the policies trained without dynamics randomization, implying that this was a useful quantity to encode into memory.}
\label{table:com_error}
\end{table}


\section{Conclusion} 

In this work, we showed that we were able to successfully and consistently train recurrent neural networks to control a Cassie robot. The policies were learned and tested first in simulation, then transferred to the robot, demonstrating the robustness and promise of this approach. We also demonstrate that recurrent neural networks appear to require some type of dynamics randomization to consistently transfer to hardware for our task, and offer the explanation that recurrent neural networks tend to overfit the dynamics of the simulation. We further show that recurrent networks are capable of encoding important information about disturbances in the dynamics into their memory, and that this information can be extracted using a decoder network; thus, recurrent networks can perform, for our task, online system identification.

Our success in these areas has implications for those seeking to apply reinforcement learning to complex dynamical systems. The ability of recurrent neural networks to do online system identification as well as behave as controllers means that they have the potential to reduce model complexity while increasing robustness. Future work might include investigating how drastic disturbances must be during training for a recurrent neural network to account for them rather than just being robust to them. Further investigation could also look into the limits of what a recurrent network is able to infer from dynamics. RNNs could also give insight into which aspects of dynamical systems are highly sensitive so users can decide which control areas require more focus than others. We feel that our findings relating to memory-based control here are not only interesting for practical reasons, but also consistent with the motivation behind the creation and use of disturbance observers in classical control methods.

\section*{Acknowledgements}
This work was partially supported by NSF grant IIS-1849343 and DARPA contract W911NF-16-1-0002. Thanks to Kevin Green, Yesh Godse, and Pedro Morais for insightful discussions, support, and feedback regarding this project. Thanks to Stephen Offer and Intel for providing access to the vLab computing cluster.

\bibliographystyle{unsrtnat}
\bibliography{references}

\begin{thebibliography}{25}
\providecommand{\natexlab}[1]{#1}
\providecommand{\url}[1]{\texttt{#1}}
\expandafter\ifx\csname urlstyle\endcsname\relax
  \providecommand{\doi}[1]{doi: #1}\else
  \providecommand{\doi}{doi: \begingroup \urlstyle{rm}\Url}\fi

\bibitem[{Peng} et~al.(2018){Peng}, {Andrychowicz}, {Zaremba}, and
  {Abbeel}]{peng2017sim2real}
X.~B. {Peng}, M.~{Andrychowicz}, W.~{Zaremba}, and P.~{Abbeel}.
\newblock \href{https://ieeexplore.ieee.org/document/8460528}{Sim-to-Real
  Transfer of Robotic Control with Dynamics Randomization}.
\newblock In \emph{2018 IEEE International Conference on Robotics and
  Automation (ICRA)}, pages 3803--3810, May 2018.
\newblock \doi{10.1109/ICRA.2018.8460528}.
\newblock URL \url{https://ieeexplore.ieee.org/document/8460528}.

\bibitem[OpenAI et~al.(2019)OpenAI, Akkaya, Andrychowicz, Chociej, Litwin,
  McGrew, Petron, Paino, Plappert, Powell, Ribas, Schneider, Tezak, Tworek,
  Welinder, Weng, Yuan, Zaremba, and Zhang]{openai2019rubikscube}
OpenAI, Ilge Akkaya, Marcin Andrychowicz, Maciek Chociej, Mateusz Litwin, Bob
  McGrew, Arthur Petron, Alex Paino, Matthias Plappert, Glenn Powell, Raphael
  Ribas, Jonas Schneider, Nikolas Tezak, Jerry Tworek, Peter Welinder, Lilian
  Weng, Qiming Yuan, Wojciech Zaremba, and Lei Zhang.
\newblock \href{https://arxiv.org/abs/1910.07113}{Solving Rubik's Cube with a
  Robot Hand}.
\newblock \emph{arXiv preprint}, 2019.
\newblock URL \url{https://arxiv.org/abs/1910.07113}.

\bibitem[Andrychowicz et~al.(2020)Andrychowicz, Baker, Chociej, Józefowicz,
  McGrew, Pachocki, Petron, Plappert, Powell, Ray, Schneider, Sidor, Tobin,
  Welinder, Weng, and Zaremba]{openai2018manipulation}
OpenAI:~Marcin Andrychowicz, Bowen Baker, Maciek Chociej, Rafal Józefowicz,
  Bob McGrew, Jakub Pachocki, Arthur Petron, Matthias Plappert, Glenn Powell,
  Alex Ray, Jonas Schneider, Szymon Sidor, Josh Tobin, Peter Welinder, Lilian
  Weng, and Wojciech Zaremba.
\newblock \href{https://doi.org/10.1177/0278364919887447}{Learning dexterous
  in-hand manipulation}.
\newblock \emph{The International Journal of Robotics Research}, 39\penalty0
  (1):\penalty0 3--20, 2020.
\newblock \doi{10.1177/0278364919887447}.
\newblock URL \url{https://doi.org/10.1177/0278364919887447}.

\bibitem[Xie et~al.(2019)Xie, Clary, Dao, Morais, Hurst, and van~de
  Panne]{xie2019iterative}
Zhaoming Xie, Patrick Clary, Jeremy Dao, Pedro Morais, Jonathan Hurst, and
  Michiel van~de Panne.
\newblock
  \href{https://zhaomingxie.github.io/publications/corl_2019.pdf}{Learning
  Locomotion Skills for Cassie: Iterative Design and Sim-to-Real}.
\newblock In \emph{3rd Conference on Robotic Learning (CORL)}, 2019.
\newblock URL \url{https://zhaomingxie.github.io/publications/corl_2019.pdf}.

\bibitem[Tan et~al.(2018)Tan, Zhang, Coumans, Iscen, Bai, Hafner, Bohez, and
  Vanhoucke]{tan2018simtoreal}
Jie Tan, Tingnan Zhang, Erwin Coumans, Atil Iscen, Yunfei Bai, Danijar Hafner,
  Steven Bohez, and Vincent Vanhoucke.
\newblock \href{http://www.roboticsproceedings.org/rss14/p10.html}{Sim-to-Real:
  Learning Agile Locomotion For Quadruped Robots}.
\newblock In \emph{Proceedings of Robotics: Science and Systems}, Pittsburgh,
  Pennsylvania, June 2018.
\newblock \doi{10.15607/RSS.2018.XIV.010}.
\newblock URL \url{http://www.roboticsproceedings.org/rss14/p10.html}.

\bibitem[Hwangbo et~al.(2019)Hwangbo, Lee, Dosovitskiy, Bellicoso, Tsounis,
  Koltun, and Hutter]{hwangbo2019learning}
Jemin Hwangbo, Joonho Lee, Alexey Dosovitskiy, Dario Bellicoso, Vassilios
  Tsounis, Vladlen Koltun, and Marco Hutter.
\newblock Learning agile and dynamic motor skills for legged robots.
\newblock \emph{Science Robotics}, 4\penalty0 (26):\penalty0 eaau5872, 2019.

\bibitem[Koul et~al.(2019)Koul, Fern, and Greydanus]{koul2018fsrnn}
Anurag Koul, Alan Fern, and Sam Greydanus.
\newblock
  \href{https://iclr.cc/Conferences/2019/Schedule?showEvent=1062}{Learning
  Finite State Representations of Recurrent Policy Networks}.
\newblock In \emph{International Conference on Learning Representations}, 2019.
\newblock URL \url{https://iclr.cc/Conferences/2019/Schedule?showEvent=1062}.

\bibitem[Arras et~al.(2017)Arras, Montavon, M{\"u}ller, and
  Samek]{arras2017RNNpredictions}
Leila Arras, Gr{\'e}goire Montavon, Klaus-Robert M{\"u}ller, and Wojciech
  Samek.
\newblock \href{https://www.aclweb.org/anthology/W17-5221}{Explaining Recurrent
  Neural Network Predictions in Sentiment Analysis}.
\newblock In \emph{Proceedings of the 8th Workshop on Computational Approaches
  to Subjectivity, Sentiment and Social Media Analysis}, pages 159--168,
  Copenhagen, Denmark, September 2017. Association for Computational
  Linguistics.
\newblock \doi{10.18653/v1/W17-5221}.
\newblock URL \url{https://www.aclweb.org/anthology/W17-5221}.

\bibitem[Cechin et~al.(2003)Cechin, Regina, Simon, and Stertz]{cechin2003state}
Adelmo~Luis Cechin, D~Regina, P~Simon, and K~Stertz.
\newblock \href{https://ieeexplore.ieee.org/document/1245447}{State automata
  extraction from recurrent neural nets using k-means and fuzzy clustering}.
\newblock In \emph{23rd International Conference of the Chilean Computer
  Science Society, 2003. SCCC 2003. Proceedings.}, pages 73--78. IEEE, 2003.
\newblock URL \url{https://ieeexplore.ieee.org/document/1245447}.

\bibitem[Wierstra et~al.(2010)Wierstra, F{\"o}rster, Peters, and
  Schmidhuber]{wierstra2010recurrent}
Daan Wierstra, Alexander F{\"o}rster, Jan Peters, and J{\"u}rgen Schmidhuber.
\newblock
  \href{https://academic.oup.com/jigpal/article/18/5/620/751594}{Recurrent
  policy gradients}.
\newblock \emph{Logic Journal of the IGPL}, 18\penalty0 (5):\penalty0 620--634,
  2010.
\newblock URL \url{https://academic.oup.com/jigpal/article/18/5/620/751594}.

\bibitem[Vukobratovic and Borovac(2004)]{zmp2004}
Miomir Vukobratovic and Branislav Borovac.
\newblock
  \href{https://www.worldscientific.com/doi/abs/10.1142/S0219843604000083}{Zero-Moment
  Point - Thirty Five Years of its Life.}
\newblock \emph{I. J. Humanoid Robotics}, 1:\penalty0 157--173, 03 2004.
\newblock \doi{10.1142/S0219843604000083}.
\newblock URL
  \url{https://www.worldscientific.com/doi/abs/10.1142/S0219843604000083}.

\bibitem[Westervelt et~al.(2003)Westervelt, Grizzle, and
  Koditschek]{westervelt2003hybrid}
Eric~R Westervelt, Jessy~W Grizzle, and Daniel~E Koditschek.
\newblock \href{https://ieeexplore.ieee.org/document/1166523}{Hybrid zero
  dynamics of planar biped walkers}.
\newblock \emph{IEEE transactions on automatic control}, 48\penalty0
  (1):\penalty0 42--56, 2003.
\newblock URL \url{https://ieeexplore.ieee.org/document/1166523}.

\bibitem[Rezazadeh et~al.(2015)Rezazadeh, Hubicki, Jones, Peekema, Van~Why,
  Abate, and Hurst]{atrias2015}
Siavash Rezazadeh, Christian Hubicki, Mikhail Jones, Andrew Peekema, Johnathan
  Van~Why, Andy Abate, and Jonathan Hurst.
\newblock \href{https://doi.org/10.1115/DSCC2015-9899}{Spring-Mass Walking With
  ATRIAS in 3D: Robust Gait Control Spanning Zero to 4.3 KPH on a Heavily
  Underactuated Bipedal Robot}.
\newblock Dynamic Systems and Control Conference, 10 2015.
\newblock \doi{10.1115/DSCC2015-9899}.
\newblock URL \url{https://doi.org/10.1115/DSCC2015-9899}.
\newblock V001T04A003.

\bibitem[Kim et~al.(2018)Kim, Kim, Kim, Sim, and Park]{kim2018disturbance}
Mingon Kim, Jung~Hoon Kim, Sanghyun Kim, Jaehoon Sim, and Jaeheung Park.
\newblock \href{https://ieeexplore.ieee.org/document/8460618}{Disturbance
  observer based linear feedback controller for compliant motion of humanoid
  robot}.
\newblock In \emph{2018 IEEE International Conference on Robotics and
  Automation (ICRA)}, pages 403--410. IEEE, 2018.
\newblock URL \url{https://ieeexplore.ieee.org/document/8460618}.

\bibitem[Paine et~al.(2015)Paine, Mehling, Holley, Radford, Johnson, Fok, and
  Sentis]{paine2015actuator}
Nicholas Paine, Joshua~S Mehling, James Holley, Nicolaus~A Radford, Gwendolyn
  Johnson, Chien-Liang Fok, and Luis Sentis.
\newblock
  \href{https://onlinelibrary.wiley.com/doi/abs/10.1002/rob.21556}{Actuator
  control for the NASA-JSC valkyrie humanoid robot: A decoupled dynamics
  approach for torque control of series elastic robots}.
\newblock \emph{Journal of Field Robotics}, 32\penalty0 (3):\penalty0 378--396,
  2015.
\newblock URL \url{https://onlinelibrary.wiley.com/doi/abs/10.1002/rob.21556}.

\bibitem[Heess et~al.(2015)Heess, Hunt, Lillicrap, and
  Silver]{Heess2015memorybased}
Nicolas Heess, Jonathan~J. Hunt, Timothy~P. Lillicrap, and David Silver.
\newblock \href{https://arxiv.org/abs/1512.04455}{Memory-based control with
  recurrent neural networks}.
\newblock \emph{CoRR}, abs/1512.04455, 2015.
\newblock URL \url{http://arxiv.org/abs/1512.04455}.

\bibitem[{Boeing} and {Bräunl}(2012)]{boeing2012leveraging}
A.~{Boeing} and T.~{Bräunl}.
\newblock \href{https://ieeexplore.ieee.org/document/6485313}{Leveraging
  multiple simulators for crossing the reality gap}.
\newblock In \emph{2012 12th International Conference on Control Automation
  Robotics Vision (ICARCV)}, pages 1113--1119, Dec 2012.
\newblock URL \url{https://ieeexplore.ieee.org/document/6485313}.

\bibitem[Sutton and Barto(2018)]{Sutton1998}
Richard~S. Sutton and Andrew~G. Barto.
\newblock
  \emph{\href{http://incompleteideas.net/book/the-book-2nd.html}{Reinforcement
  Learning: An Introduction}}.
\newblock The MIT Press, second edition, 2018.
\newblock URL \url{http://incompleteideas.net/book/the-book-2nd.html}.

\bibitem[Schulman et~al.(2017)Schulman, Wolski, Dhariwal, Radford, and
  Klimov]{schulman2017proximal}
John Schulman, Filip Wolski, Prafulla Dhariwal, Alec Radford, and Oleg Klimov.
\newblock \href{https://arxiv.org/abs/1707.06347}{Proximal Policy Optimization
  Algorithms}, 2017.
\newblock URL \url{https://arxiv.org/abs/1707.06347}.

\bibitem[{Singla} et~al.(2019){Singla}, {Padakandla}, and
  {Bhatnagar}]{UAVmemory}
A.~{Singla}, S.~{Padakandla}, and S.~{Bhatnagar}.
\newblock \href{https://ieeexplore.ieee.org/document/8917687}{Memory-Based Deep
  Reinforcement Learning for Obstacle Avoidance in UAV With Limited Environment
  Knowledge}.
\newblock \emph{IEEE Transactions on Intelligent Transportation Systems}, pages
  1--12, 2019.
\newblock ISSN 1558-0016.
\newblock \doi{10.1109/TITS.2019.2954952}.
\newblock URL \url{https://ieeexplore.ieee.org/document/8917687}.

\bibitem[Hausknecht and Stone(2015)]{recurrentDQN}
Matthew Hausknecht and Peter Stone.
\newblock
  \href{http://www.cs.utexas.edu/~pstone/Papers/bib2html/b2hd-SDMIA15-Hausknecht.html}{Deep
  Recurrent Q-Learning for Partially Observable MDPs}.
\newblock In \emph{AAAI Fall Symposium on Sequential Decision Making for
  Intelligent Agents (AAAI-SDMIA15)}, November 2015.
\newblock URL
  \url{http://www.cs.utexas.edu/~pstone/Papers/bib2html/b2hd-SDMIA15-Hausknecht.html}.

\bibitem[Hochreiter and Schmidhuber(1997)]{schmidhuber1997lstm}
Sepp Hochreiter and J\"{u}rgen Schmidhuber.
\newblock
  \href{https://www.mitpressjournals.org/doi/10.1162/neco.1997.9.8.1735}{Long
  Short-Term Memory}.
\newblock \emph{Neural Comput.}, 9\penalty0 (8):\penalty0 1735–1780, November
  1997.
\newblock ISSN 0899-7667.
\newblock \doi{10.1162/neco.1997.9.8.1735}.
\newblock URL \url{https://doi.org/10.1162/neco.1997.9.8.1735}.

\bibitem[Yu et~al.(2017)Yu, Tan, Liu, and Turk]{yu2017sysid}
Wenhao Yu, Jie Tan, C.~Karen Liu, and Greg Turk.
\newblock \href{http://www.roboticsproceedings.org/rss13/p48.html}{Preparing
  for the Unknown: Learning a Universal Policy with Online System
  Identification}.
\newblock In \emph{Proceedings of Robotics: Science and Systems}, Cambridge,
  Massachusetts, July 2017.
\newblock \doi{10.15607/RSS.2017.XIII.048}.
\newblock URL \url{http://www.roboticsproceedings.org/rss13/p48.html}.

\bibitem[Paszke et~al.(2019)Paszke, Gross, Massa, Lerer, Bradbury, Chanan,
  Killeen, Lin, Gimelshein, Antiga, Desmaison, Kopf, Yang, DeVito, Raison,
  Tejani, Chilamkurthy, Steiner, Fang, Bai, and Chintala]{pytorch}
Adam Paszke, Sam Gross, Francisco Massa, Adam Lerer, James Bradbury, Gregory
  Chanan, Trevor Killeen, Zeming Lin, Natalia Gimelshein, Luca Antiga, Alban
  Desmaison, Andreas Kopf, Edward Yang, Zachary DeVito, Martin Raison, Alykhan
  Tejani, Sasank Chilamkurthy, Benoit Steiner, Lu~Fang, Junjie Bai, and Soumith
  Chintala.
\newblock
  \href{https://papers.nips.cc/paper/9015-pytorch-an-imperative-style-high-performance-deep-learning-library}{PyTorch:
  An Imperative Style, High-Performance Deep Learning Library}.
\newblock In H.~Wallach, H.~Larochelle, A.~Beygelzimer, F.~d'Alch\'{e} Buc,
  E.~Fox, and R.~Garnett, editors, \emph{Advances in Neural Information
  Processing Systems 32}, pages 8024--8035. Curran Associates, Inc., 2019.
\newblock URL
  \url{https://papers.nips.cc/paper/9015-pytorch-an-imperative-style-high-performance-
  deep-learning-library}.

\bibitem[Kingma and Ba(2014)]{kingma2014adam}
Diederik~P Kingma and Jimmy Ba.
\newblock \href{https://arxiv.org/abs/1412.6980}{Adam: A method for stochastic
  optimization}.
\newblock \emph{arXiv preprint arXiv:1412.6980}, 2014.
\newblock URL \url{https://arxiv.org/abs/1412.6980}.

\end{thebibliography}

\end{document}